\documentclass{article} 
\usepackage{iclr2020_conference,times}
\usepackage[utf8]{inputenc}

\usepackage{amsmath,amsfonts,bm}

\def\eqref#1{equation~\ref{#1}}

\def\1{\bm{1}}

\DeclareMathAlphabet{\mathsfit}{\encodingdefault}{\sfdefault}{m}{sl}
\SetMathAlphabet{\mathsfit}{bold}{\encodingdefault}{\sfdefault}{bx}{n}

\usepackage{hyperref}
\usepackage{url}
\usepackage{graphicx}
\usepackage{comment}

\title{A Deep Learning approach for determining effects of \textit{TUTA ABSOLUTA} in Tomato plants}

\author{\textbf{Denis P. Rubanga } \textsuperscript{*1}, \textbf{ Loyani K. Loyani } \textsuperscript{2}, \textbf{ Mgaya Richard} \textsuperscript{3}  \textbf{{ \&} Sawahiko Shimada} \textsuperscript{1}\\
\textsuperscript{1} Department \ of \ Agricultural \ Engineering \\ 
Tokyo University of Agriculture, Tokyo, Japan \\
\textsuperscript{2}
School of Computational {\&} Communication Science {\&} Technology {\&} Engineering,\\
The Nelson Mandela African Institution of Science {\&} Technology, Arusha, Tanzania.\\
\textsuperscript{3}Department of Engineering Sciences {\&} Technology,\\
Sokoine University Agriculture, Morogoro, Tanzania.\\
\textsuperscript{*}\texttt{denispastoty@gmail.com} 
}

\iclrfinalcopy 

\begin{document}

\maketitle

\begin{abstract}
Early quantification of \textit{Tuta absoluta} pest’s effects in tomato plants is a very important factor in controlling and preventing serious damages of the pest. The invasion of \textit{Tuta absoluta} is considered a major threat to tomato production causing heavy loss ranging from 80 to 100 percent when not properly managed. Therefore, real-time and early quantification of tomato leaf miner \textit{Tuta absoluta}, can play an important role in addressing the issue of pest management and enhance farmers’ decisions. In this study, we propose a Convolutional Neural Network (CNN)  approach in determining the effects of \textit{Tuta absoluta} in tomato plants. Four CNN pretrained architectures (VGG16, VGG19, ResNet and Inception-V3) were used in training classifiers on a dataset containing health and infested tomato leaves collected from real field experiments. Among the pretrained architectures, experimental results showed that Inception-V3 yielded the best results with an average accuracy of 87.2 percent in estimating the severity status of \textit{Tuta absoluta} in tomato plants. The pretrained models could also easily identify High Tuta severity status compared to other severity status (Low tuta and No tuta)
\end{abstract}

\section{Introduction}
\label{intro}
The current world population is expected to reach 9.8 billion in 2050 \citep{UN:2017}. To ensure global food security and meet the future demand for high quantity and quality food, the agricultural industry must be much efficient and robust.
Sub-saharan small scale farmers rely on tomatoes to earn income. However, tomato productivity is threatened by the invasion of an exotic pest known as tomato leaf miner \textit{Tuta absoluta} \citep{zekeya2017tomato}. The pest has become a major drawback to tomato production causing heavy losses in tomato produce ranging from 80\% to 100\%  \citep{desneux2011invasive, chidege2016first} (See Figure.1). Since 2008, the pest has invaded and spread to 75\% of African countries causing huge economic losses \citep{guimapi2016modeling}. Nevertheless, the extension service to provide farmers with appropriate knowledge about plant disease and pest managements are limited \citep{maginga2018extension}.

\begin{figure}[h]
    \begin{center}
    \includegraphics[width=0.9\textwidth]{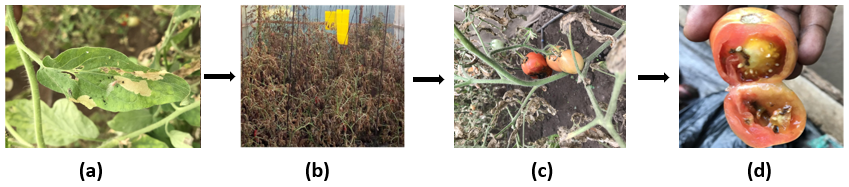}
    \caption{Larvae, the most dangerous stage of \textit{Tuta absoluta}’s life cycle. (a) Tomato leaf with \textit{T.absoluta} mine (b) \textit{T.absoluta} severe damage on our in-house tomato field (c) Affected tomato fruits(d) Damaged tomato fruit on market.}
    \label{fig:larvaeffects}
    \end{center}
\end{figure}

Despite existence of various ways of controlling tomato pests, there has not been an efficient mechanism to determine the severity of \textit{T.absoluta}’s effects at early stages before causing great yield loss to the farmers. Inspired by the advancement and promising results of deep learning techniques in image-based plant pest and disease recognition, this research proposes the use of Convolutional Neural Network (CNN) model to determine the severity status of \textit{T.absoluta}’s damage on tomato plants at early stage of tomato growth. This will enhance farmers’ intelligently informed decisions in controlling the pest and improve tomato productivity in order to rescue farmers from losses they incur every year.

\section{Related Works}
\label{literature}
The advances in computer vision and machine learning techniques such as deep learning and specifically Convolutional Neural Networks (CNN) have presented promising and impressive results in tasks such as identification and  classification of a diverse range of plant diseases and pests \citep{singh2016machine}. For instance, \citet{brahimi2017deep} presented deep models (AlexNet and GoogleNet), trained using a large dataset of 14,828 images to identify 9 tomato diseases. 
Also, \citet{ferentinos2018deep} used several deep models (AlexNet, GoogLeNet and VGG) in recognizing 58 diseases from a dataset of 87,848 leaf images of different plants from PlantVillage repository \citep{hughes2015open}.
\citet{zhang2018can} proposed pretrained CNN models to identify 8 tomato diseases from an open access repository of 5550 images. \citet{liang2019pd2se} proposed a multitasking system consisting of ResNet50 architecture capable to diagnose diseases, recognizing the plant species and estimating the severity of diseases using PlantVillage dataset \citep{hughes2015open}. Other works include automatic and multi-task systems based on CNN for classification task \citep{esgario2020deep, wang2017automatic}.

The aforementioned works address plant disease problems. However, few works have focused on estimating plant pest stress severity challenges. Plant stress severity have been limited to plant diseases. Some of these works such as \citet{wang2017automatic}, \citet{brahimi2017deep} and \citet{ferentinos2018deep} used images from online repositories, that do not reflect the real-life situation and put these model’s performance into questions that could limit their applicability in real field situations. 

Our research, contributes to the very few works on estimating tomato plant pest severity. To the best of our knowledge, this first novel work proposes approaches in determining severity status of \textit{T.absoluta}’s effects on tomato plants.
With the lack of image data, we established a data collection strategy and collected our own dataset from the real field. We propose a deep learning-based approach for determining the effects of \textit{T.absoluta} at early stages of tomato plant’s growth. The study will help farmers and extension officers to make intelligently informed decisions that could improve tomato productivity and rescue farmers from the losses they incur every year.
\section{Material and Methods}
\label{methodology}

\subsection{Datasets}
\label{data}
Four (4) In-house data collection works were conducted in two of the major areas prone to \textit{T.absoluta} infestation (Arusha and Morogoro - Tanzania) as summarized in Table. 1. The table shows factors that were put into consideration to have a vast diverse dataset of the real field situations i.e regions of the country that are highly infestated with \textit{T.absoluta}, crop cycle season, mainly grown tomato varieties and mainly practiced farming systems. 
We planted healthy tomato seedlings (free from other diseases and pests), inoculated some plants on a range of 2 to 8 \textit{T.absoluta} larvae per plant on the second day after transplanting and on a daily basis took pictures of every plant between 08:00 and 10:00 A.M consecutively for two weeks.  
For this work, we only picked 1384 \textit{T.absoluta} infested plant images, separated them into two categories of \textit{T. absoluta} damage severity status examined by agricultural expert as Low \textit{Tuta} (plants with less than 3 T. absoluta) and High \textit{Tuta} (more than 3 \textit{T. absoluta}). A total of 692 images of Low \textit{Tuta}, 692 of High \textit{Tuta} images and 2768 images of No \textit{Tuta} from the whole dataset was used, finally making three classes i.e No \textit{Tuta}, low \textit{Tuta} and high \textit{Tuta} as shown in Figure 2.

\begin{table}[h]
\caption{Data collection set-up and factors considered for each experiment}
\label{sample-table}
\begin{center}
\begin{tabular}{llllll} 
\hline \\ 
\multicolumn{1}{c}{\bf duration}&
\multicolumn{1}{c}{\bf season}&
\multicolumn{1}{c}{\bf region}&
\multicolumn{1}{c}{\bf variety}&
\multicolumn{1}{c}{\bf farming system}&
\multicolumn{1}{c}{\bf images}
\\ \hline \\ 
Aug - Nov 2018 & dry & north & 1 & drip, furrow, bund & 2248 \\
Jan - May 2018 & dry & north & 3 & drip & 2012 \\
Oct - Dec 2019 & dry/wet & north & 3 & drip & 4060 \\
Jan - Apr 2020 & wet & east & 2 & drip, furrow, bund & 2916 \\
 \hline \\ 

\end{tabular}
\end{center}
\end{table}

To reduce the bias due to imbalance data (our dataset has more No \textit{Tuta} images), 10\% of the images were held as test set, and the remaining 90\% were sub-divided into training and validation sets in the ratio of 85:15. Also, No \textit{Tuta} images were divided into 4 clusters of images while retaining 10\% for testing. Therefore making four datasets each with a total of 1623 for training, 230 images for validation and 218 for testing. 
\begin{figure}[!ht]
    \centering
    \includegraphics[width=0.85\textwidth]{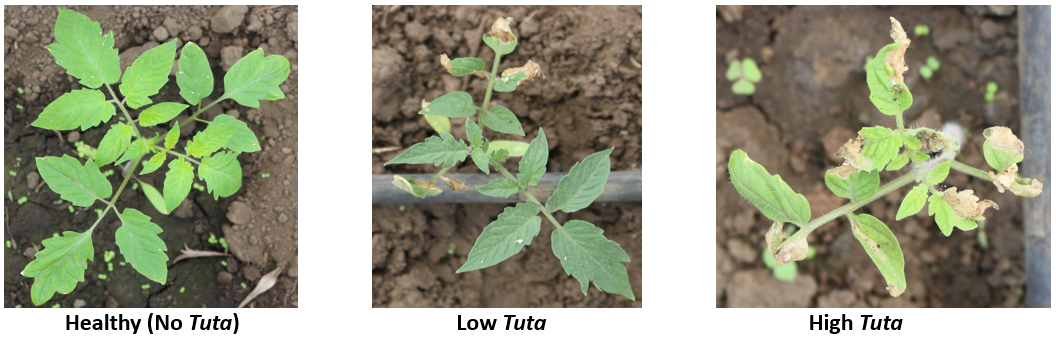}
    \caption{Some images collected from field showing damage status of \textit{T. absoluta}.}
    \label{tutastatus}
\end{figure}

\subsection*{Methods}
\label{methods}
The main target of this work was to be able to identify \textit{T.absoluta} damaged tomato plant severity status that could help to make clear distinction between the three classes. We therefore, choose four CNN architectures; VGG16, VGG19 \citep{simonyan2014very}, ResNet50 and Inception-V3\citep{szegedy2016rethinking} to train classifiers on our dataset containing the three tomato severity status. 


\subsection*{Image Preprocessing.}
To increase image number and reduce the variation within each image for \textit{T.absoluta} severity status classification, augmentations were performed on both the training and validation. All images were first resized to 256 x 256 pixels for VGG16, VGG19 and ResNet50 and 384 x 384 pixel for Inception-V3 and randomly augmented for each epoch of training. Each image was also randomly rotated in the range of (-360,+360), degrees also randomly sheared in the range of 0.3 To account for illumination variance, pixel intensity was randomly shifted within the range (-25, +25), shifting all colour channels uniformly. In addition, pixel intensity was randomly scaled within the range of (0.75, 1.25). The images were also zoomed within (0.5,1) range. Finally, the images were flipped horizontally, cropped back to 224 X 224 pixels for VGG16, VGG19 and ResNet50 for 299 X 299 pixels required for architecture's input layer \citep{perez2017effectiveness}


\subsection*{Training our classifier} 
We used four ImageNet \citep{deng2009imagenet} pretrained architectures VGG16, VGG19, ResNet50 and InceptionV3 as classifiers. The fully connected layer for each pretrained architecture was replaced by the new layer (3-class classifier for our dataset). 
We trained our classifiers using 50 epochs with a batch size of 16 and using Keras \citep{kerasgit} implementation of Adam \citep{kingma2014adam}, a first-order gradient-based method for stochastic optimization. The initial learning-rate (lr) was set to lr=10e4, and was halved every time the validation loss did not decrease after 32 epochs in batches of 16 images, and aborted if the validation loss did not decrease after 32 epochs. The model with the smallest running validation loss was continuously saved, in order to re-start the training after an abortion. In such cases, training was repeated with the initial learning rate lr=0.5x10e4. With the four subset dataset, we run all the four models on each of the subsets. 

\subsection*{Implementation} The experiments were performed on Ubuntu workstation, pre-installed with Ubuntu 18.04 equipped with one Intel Core i9-9900K 3.6 GHz CPU (64 Gb RAM) accelerated by one GeForce RTX 2080Ti Graphical Processing Unit (GPU) (12 GB memory). We trained 50 epoch for each model and it took an average of 41 minute on a complete model training powered by Keras deep learning library using Tensorflow \citep{abadi2016tensorflow} as backend. In total, about 13 hours were required to run training on the 16 runs of the models i.e 4 runs for each of the 4 CNN architecture.

\section*{Results and Discussion}
We used evaluation metrics F1--score, precision and recall accuracy and the overall evaluation metrics was a result of averaging over the 4 runs on each dataset of each CNN architecture as summarized in table. 2.\\
The main goal was severity status of \textit{T.absoluta} determination. In term determining the severity status images, all models precision accuracy was highest in identifying High Tuta images i.e 90.5\%, 90.5\%, 90.3\% and 91.5\%. VGG16 and Inception-V3 had the highest recall accuracy i.e 96.5\% on No Tuta images. Also all models had F1-score highest for High Tuta images. All the four models had the lowest evaluation metrics accuracy in determining Low Tuta images. Among the trained models, Inception-V3 model had the highest accuracy of 87.2\% on the test set.

\begin{table}[h]
\caption{Four pretrained model evaluation metrics accuracy precision (PRC), recall (RCL), F1--score (F1-S) accuracy and Overall average accuracy and loss on testing datasset.}
\label{sample-table}
\begin{center}
\resizebox{\textwidth}{!}{
\begin{tabular}{llllllllllllllll}
\hline &
\multicolumn{3}{c}{\bf VGG16}&
\multicolumn{1}{c}{\bf}& 
\multicolumn{3}{c}{\bf VGG19}&
\multicolumn{1}{c}{\bf}& 
\multicolumn{3}{c}{\bf ResNet50}&
\multicolumn{1}{c}{\bf}& 
\multicolumn{3}{c}{\bf Inception-V3}\\
{\bf Severity}&PRC&RCL&FI-S&&PRC&RCL&FI-S&&PRC&RCL&FI-S&&PRC&RCL&FI-S\\
 \hline
No Tuta & 
0.877&\textbf{0.965}&0.918&&0.883&0.918&0.918&&0.878&0.900&0.890&&0.895&\textbf{0.933}&0.915\\
Low Tuta & 
0.760&0.355&0.448&&0.708&0.538&0.595&&0.500&0.445&0.470&&0.660&0.518&0.575\\
High Tuta & 
\textbf{0.905}&0.940&\textbf{0.920}&& \textbf{0.905}&\textbf{0.948}& \textbf{0.920}&&\textbf{0.903}&\textbf{0.910}&\textbf{0.905}&&\textbf{0.915}&0.930&\textbf{0.923} \\
 \hline 
Average Accuracy & 
0.871&&&&0.783&&&&0.837&&&&\textbf{0.872}&\\
Loss & 
\textbf{0.152}&&&&\textbf{0.258}&&&&\textbf{0.334}&&&&0.205 \\
\hline 
\end{tabular}}
\end{center}
\end{table}

\section{Conclusion and Future Work}
This paper proposes pretrained deep
learning models  for determining severity status of \textit{ T.Absoluta} tomato damages plants. For the accomplishment of this work, we used images containing health and \textit{T.Absoluta} infested tomato plant images collected from in-house experiments. Among the pretrained models, we showed that Inception-V3 model performed best, achieving an averaged accuracy of 87.2\% on the test set compared to other models.\\
Among, the three severity status, all models could more easily identify High tuta images than other severity status based on the evaluation metrics. The comparison of the evaluation metrics on each of the severity status reveals that it is a bit harder to detect Low Tuta than High Tuta and No Tuta images. 
With the goal of early identification of \textit{T.Absoluta} severity status in tomato plants, we clearly show the success of using deploying CNN models in such tasks. High Tuta severity status being determined as early as the first two weeks of plant growth cycle is important to reduce severe loss that are accounted when no preventive and management practices are not done. 

In future work, we intend to experiment on other CNN based models for task such as instance segmentation for localization of \textit{T.Absoluta} images.Infact, ongoing work includes, annotation of images at infested plant based and localised \textit{T.Absoluta} patches on the plant leaves for instance segmentation tasks.


\bibliography{iclr2020_conference}
\bibliographystyle{iclr2020_conference}

\end{document}